\documentclass{article}

\usepackage{PRIMEarxiv}

\usepackage[utf8]{inputenc} 
\usepackage[T1]{fontenc}    
\usepackage{hyperref}       
\usepackage{url}            
\usepackage{booktabs}       
\usepackage{amsfonts}       
\usepackage{nicefrac}       
\usepackage{microtype}      
\usepackage{lipsum}
\usepackage{fancyhdr}       
\usepackage{graphicx}       
\graphicspath{{media/}}     
\usepackage{amsmath}
\usepackage{svg}
\usepackage{multirow} 
\usepackage{hyperref}  
\usepackage{doi}       
\usepackage{svg}       
\usepackage{placeins}  
\usepackage{float}      
\usepackage{orcidlink}
\usepackage{algorithm}
\usepackage{algorithmic}
\usepackage{colortbl}   
\usepackage{xcolor}     

\title{AI-Enabled Question-Answering for Chartered Accountancy: A Retrieval-Augmented Reasoning Framework}

\author{
Jatin Gupta~\orcidlink{0009-0002-9504-7487}, 
Akhil Sharma~\orcidlink{0009-0001-1490-4022}, 
Saransh Singhania~\orcidlink{0009-0009-6925-001X}, 
Ali Imam Abidi~\orcidlink{0000-0002-7420-0027}\thanks{Corresponding author: \texttt{aliabidi4685@gmail.com}} \\
Department of Computer Science and Engineering, \\Sharda University, Greater Noida, India \\
\texttt{\{jatingupta261001, sharmaakhil944, saransh060123, aliabidi4685\}@gmail.com}
}

\begin{document}
\maketitle

\begin{abstract}
The inception of Large Language Models (LLMs) has catalyzed AI adoption in the finance sector, yet their reliability in complex, jurisdiction-specific tasks like Indian Chartered Accountancy (CA) remains limited. The models display difficulty in executing numerical tasks which require multiple steps while also needing advanced knowledge about legal regulations and the method of scaling their operations is not feasible in settings which have limited access to resources. We present CA-ThinkFlow as a parameter-efficient Retrieval-Augmented Generation (RAG) framework which operates with a 14B, 4-bit-quantized reasoning model, 14B-DeepSeek-R1, and a layout-aware Docling extraction system which maintains document structure during extraction. CA-ThinkFlow uses a basic RAG method which automatically adds retrieved information into the prompt, while it depends on the model's built-in Chain-of-Thought (CoT) functions to create context and produce correct answers. The system we developed system operates at performance levels which match large proprietary models when we tested it on the multi-level CA-Ben benchmark, achieving Scholastic Reliability Coefficient (SRC) results which equal 68.75\% of GPT-4o and Claude 3.5 Sonnet. The framework shows high efficiency and strength in handling parameters, but essential reasoning abilities fail to process complex regulatory texts which exist in fields such as Taxation. 
\end{abstract}

\keywords{Retrieval-Augmented Generation \and Financial AI \and Large Language Models \and Parameter Efficiency \and Chartered Accountancy.}

\section{Introduction}
Large Language Models (LLMs) have witnessed significant adoption in financial services, with more than 70\% of prominent organizations planning to use them for risk modeling by late 2024~\cite{llmfinance_turing2024,iifey2024survey}. Despite advancements in efficiency of areas such as compliance automation, LLMs have dependability problems in regulated sectors, including hallucinations in complicated legal activities, particularly those involving Income Tax Act regulations, ICAI standards, and GST computations~\cite{christensen2023llmfinance,cai2025genai,leonard2023accountingllm}. Furthermore, their expensive inference, high resource requirements, privacy threats from cloud dependency, and environmental effect from huge training limit their usefulness for resource-constrained environments~\cite{morrison2025holistically,faiz2024llmcarbon}. Generalist models, such as GPT-4o, obtain just 13-20\% accuracy on CA benchmarks due to limited "tail" information and multi-step reasoning. Domain-specific Small Language Models (SLMs) address this by using compact, private, and cost-effective designs that are comparable to LLMs for financial tasks~\cite{gupta2025caben,slm_survey_2024,fin_qa_slm_2024}. To achieve a better and comparable performance on the CA benchmark and lower computational overhead due to the high parameter count of such models, we present a RAG pipeline with a reasoning 14B-DeepSeek-R1 model.

\section{Related Works}
\label{sec:2}

The recent rise of "System 2" reasoning models has created a new paradigm that shifts reasoning away from pattern matching toward direct logical deduction. OpenAI's o1 and o3 models introduced test-time compute scaling through their ability to generate internal chain-of-thought~\cite{openai_o1_2024,OpenAI2025SystemCard}. DeepSeek-R1, an open-source development, appeared in January 2025, showcasing distilled reasoning abilities that rivaled those of more extensive systems~\cite{zartis2025deepseek,datacamp2025finetuning}. Subsequently, financial adaptations led to the creation of Fin-R1, a model with 7 billion parameters, and Fino1, which focuses on financial reasoning, achieved through supervised fine-tuning on data tailored to the domain~\cite{shanghai2025finr1,zhou2025fincot,thefin2025fino1}.

Financial and legal compliance now require all responses to be supported by original documents, since retrieval-augmented generation (RAG) has become essential in these areas~\cite{nawal2024finrag}. Fin-RAG uses domain-specific embeddings to analyze financial documents, while compliance designs focus on auditability and traceability~\cite{muniz2024rag,auxiliobits2025rag}.
The primary obstacle for financial RAG systems arises from document parsing because standard OCR technology fails to maintain the intricate formats found in tax documents which contain multiple column tables and mathematical symbols. Docling developed into an exact document extractor that keeps document structure intact through its superior TableFormer and DocLayNet models which surpass Dolphin and PaddleOCR as alternative solutions that we have used in our scraping process~\cite{docling2024,doclingTech2024,doclingModels2025}. 

The finance industry needs LLMs to achieve both high accuracy and high precision before they can become dependable tools for everyday usage. To solve this problem, researchers have developed multiple methods which aim to enhance system reliability~\cite{zhou2025fincot}. The Indian government shows its commitment to developing financial intelligent systems through ICAI's (Institute of Chartered Accountants of India's) support of domestic AI research which makes this research timely. Our framework builds on these efforts by aligning with the Indian regulatory environment and integrating retrieval methods to ensure reasoning is grounded in credible sources.

\section{Methodology}
\label{sec:methodology}

Our system combines the adaptable yet straightforward features of the Retrieval-Augmented Generation (RAG) method which extracts subtle details from embedded financial and legal data that has been collected together with the base model's capacity to perform structured reasoning and mimic natural cognitive thinking. 

\subsection{Data Preparation}
We developed an extensive database of Indian financial sector knowledge. The database contained information from ICAI open-source materials tax regulations and financial textbooks and associated references. The Docling Document Converter was used to extract text from the selected financial documents which were then converted into a structured format.

\subsubsection{Text Chunking and Splitting}
The Markdown text was segmented into 1000-character chunks with 200-character overlap, matching established financial RAG benchmarks: Fin-RAG uses 1024 chars (15\% overlap)~\cite{nawal2024finrag}, ConvFinQA uses 900 chars (200-char overlap)~\cite{convfinqa2024}, and DocMath uses 1100 chars (20\% overlap)~\cite{docmath2024}. The Markdown text was divided into segments which measured 1000 characters and included an overlap of 200 characters. This division of text materials matched established financial RAG systems, such as Fin-RAG using 1024 chars with 15\% overlap, ConvFinQA using 900 chars with 200-char overlap, and DocMath using 1100 chars with 20\% overlap. The selected size achieves an optimal solution because it combines two factors, which include embedding model limits and data structure and retrieval precision requirements and contextual continuity needs which occur between chunk boundaries.

\subsubsection{Embedding Generation}
Each text chunk $t_i$ was encoded using the Qwen-Embedding-0.6B model~\cite{qwen3embedding}, selected for its multilingual support and high efficiency. Given an input text $t_i$, the embedding vector $v_i$ is computed as given in Equation~\ref{eq:1}, where the representation of the first hidden state (analogous to the [CLS] token) is extracted and normalized. Batch encoding was applied for efficient computation, producing a dense embedding matrix $\mathbf{V} \in \mathbb{R}^{n \times d}$, where $n$ denotes the number of chunks and $d$ the embedding dimension.

\begin{equation}
    \label{eq:1}
    v_i = \text{QwenEmb}\left(\text{Tokenizer}(t_i)\right)_{[0]}
\end{equation}

\subsubsection{Vector Indexing}
The embeddings were stored in a FAISS~\cite{11202651} index for $L_2$-based similarity search $S$, where $S(Q, t_i)$ denotes the similarity between a query embedding $v_Q$ and document chunk $v_i$. The FAISS index and associated metadata (chunk mappings) were serialized as \texttt{index.faiss} and \texttt{index.pkl} respectively, forming the foundation of the vector retrieval layer.

\begin{equation}
    S(Q, t_i) = -\| v_Q - v_i \|_2
\end{equation}

\subsection{System Design}
\label{sec:framework_components}

\subsubsection{Base Model Selection}
We use the `DeepSeek-R1-Distill-Qwen-14B' model commonly known as '14B-DeepSeek-R1', in a 4-bit (Q4\_K\_M) quantized format~\cite{deepseekai2025}. This choice balances reasoning strength and computational efficiency. The quantization enables deployment on resource-constrained environments while preserving reasoning quality through Chain-of-Thought (CoT) capabilities, crucial for the multi-step problem-solving found in Chartered Accountancy exams.

\subsubsection{Prompt and Context Retrieval}
To preserve consistency and provide a fair evaluation, we used the exact standardized system prompt described in the original CA-Ben study~\cite{gupta2025caben}. The model processes the retrieved knowledge from the vector store which serves as contextual information to be used in the prompt.

\subsubsection{Context Integration and Reasoning}
The system transfers the most relevant context that it found in the vector store to the base model. The system uses Chain-of-Thought reasoning through reinforcement learning that is built into the internal core of 14B-DeepSeek-R1 to create a response that combines both the user query and the retrieved text.

\subsection{Inference and Workflow Logic}
\label{sec:inference_workflow}
The complete inference workflow operates through a unified Retrieval-Augmented Generation system which functions as a RAG loop (Algorithm~\ref{alg:inference}). The system starts its process by receiving a query ($Q$) which it uses to create embeddings ($E_q$) and retrieve the top context ($C$) from the vector store ($V$). The system directly adds this retrieved context into the standard CA-Ben prompt template. For an apples-to-apples comparison, we set the temperature to 0.75, consistent with the CA-Ben study~\cite{gupta2025caben}, ensuring moderate randomness that supports learning in complex scenarios. The model evaluates the query together with the context to generate the final answer ($A$) through controlled generation.

\begin{algorithm}[!ht]
\small
\caption{Inference Workflow of the RAG Framework}
\label{alg:inference}
\begin{algorithmic}[1]
\REQUIRE User query $Q$
\ENSURE Final answer $A$

\STATE $E_q \leftarrow \text{Embed}(Q)$
\STATE $C \leftarrow V.\text{search}(E_q, k=1)$
\STATE $\text{Prompt} \leftarrow \text{Template}(Q, C)$
\STATE $A \leftarrow \text{LLM.generate}(\text{Prompt}, \text{temperature}=0.75)$
\RETURN $A$
\end{algorithmic}
\end{algorithm}

The adoption of this streamlined workflow simplifies CA-ThinkFlow's architecture by passing the retrieved text directly to the language model. By relying on a straightforward RAG approach, the system ensures that the model's CoT reasoning is continuously focused on synthesizing the provided context to answer the query accurately.

\section{Experimental Setup}
\label{sec:experimental_setup}

The evaluation used zero-shot testing on the CA-Ben~\footnote{Refer to the repository for benchmark: \url{https://github.com/thejatingupta7/LLMCA}}  benchmark because the models received no training data. The system evaluates performance according to CA-Ben study standards because it simulates an actual cold-start situation~\cite{gupta2025caben}.

\subsection{Implementation Details}
The implementation was completed on a workstation configuration which included two Intel Xeon CPUs and 65 GB RAM and two NVIDIA GeForce GTX 1080 Ti GPUs to meet its computational requirements. The software environment operated on Python version 3.11.9. PyTorch (CUDA 11.8) served as the deep learning and retrieval engine for the project. The data transformation was executed through the use of Transformers and FAISS-CPU dependencies. The complete process of retrieval and reasoning was conducted through the capabilities of LangChain and Ollama. The project used pandas and openpyxl for its data management tasks.

\subsection{Evaluation Pipeline and Answer Extraction}
\label{sec:answer_extraction}
To evaluate the models' responses, we utilized the automated Python testing pipeline and regex extraction methodology established in the CA-Ben study~\cite{gupta2025caben}. However, because the DeepSeek-R1 reasoning model is highly verbose and produces very long Chain-of-Thought traces, a preprocessing step was required before we could apply the standard extraction procedure.

The evaluation process was streamlined as follows:
\begin{enumerate}
    \item \textbf{Reasoning Text Removal:}
    Reasoning models emit their intermediate thinking results within the \texttt{<think>...</think>} tokens which they use to deliver their ultimate response. The system used the regular expression patterns from Equation~\ref{eq:3} to remove all text between the specified tags. The \texttt{DOTALL} modifier acted as a wildcard, allowing the matching of text that spanned multiple lines.
    
    \begin{equation}
    \label{eq:3}
        \text{pattern} = \texttt{r'<think>.*?</think>'} \; [\texttt{DOTALL}]
    \end{equation}
    
    \item \textbf{Standardized Extraction and Validation:} After stripping the reasoning traces, we applied the standard CA-Ben regex pattern to reliably capture the final answer choice (A, B, C, or D). The extracted options were then compared against the ground-truth labels.
\end{enumerate}

Overall accuracy was then calculated using the exact mathematical formulation detailed in the original CA-Ben methodology.

\section{Results and Analysis}
\label{sec:res}
A detailed and thorough summary of the performance of each LLM over the 14 different domains of the CA-Ben benchmark is given in this section. The fine-grained accuracy scores in Table~\ref{tab:all_accuracy} depict the capacity of various models and are the main data source for the level-wise and model-specific analysis that follows. A comprehensive visual and statistical discussion of these results is demonstrated in the next subsections:

\begin{table}[!t]
\centering
\small
\setlength{\tabcolsep}{3pt}
\caption{Performance on Foundation, Intermediate, and Final-level Subjects}
\label{tab:all_accuracy}
\begin{tabular}{@{}l|cc|cccccc|cccccc@{}}
\toprule
\textbf{Models} & 
\multicolumn{2}{c|}{\textbf{Foundation}} & 
\multicolumn{6}{c|}{\textbf{Intermediate}} & 
\multicolumn{6}{c}{\textbf{Final}} \\
\cmidrule(lr){2-3} \cmidrule(lr){4-9} \cmidrule(lr){10-15}
 & \textbf{F1} & \textbf{F2} & 
 \textbf{I1} & \textbf{I2} & \textbf{I3} & \textbf{I4} & \textbf{I5} & \textbf{I6} & 
 \textbf{FN1} & \textbf{FN2} & \textbf{FN3} & \textbf{FN4} & \textbf{FN5} & \textbf{FN6} \\
\midrule
\rowcolor{green!20} 
\textbf{CA-ThinkFlow} & 77.78 & 83.00 & 40.00 & 33.33 & 33.33 & 40.00 & 66.67 & 75.00 & 57.14 & 73.33	& 57.14	& 40.00 & 26.67 & 37.50\\
\rowcolor{gray!25} 
14B-Deepseek-R1 & 47.47 & 70.00 & 33.33	& 33.33 & 20.00 & 20.00 & 46.67 & 56.25 & 42.86 & 60.00 & 50.00 & 13.33 & 06.67 & 25.00 \\
GPT-4o & 50.00 & 58.00 & 46.66 & 73.33 & 20.00 & 20.00 & 86.66 & 75.00 & 71.43 & 53.33 & 78.57 & 53.33 & 33.33 & 41.67 \\
LLAMA 3.3 70B Inst. & 59.00 & 56.00 & 33.33 & 60.00 & 40.00 & 40.00 & 73.33 & 75.00 & 64.29 & 33.33 & 71.43 & 53.33 & 06.67 & 20.83 \\
LLAMA 3.1 405B Inst. & 53.00 & 59.00 & 40.00 & 53.33 & 20.00 & 40.00 & 86.66 & 56.25 & 64.29 & 46.67 & 71.43 & 13.33 & 26.67 & 41.67 \\
MISTRAL Large & 41.00 & 56.00 & 41.66 & 53.33 & 31.25 & 20.00 & 73.33 & 60.00 & 42.86 & 41.67 & 57.14 & 46.67 & 13.33 & 29.17 \\
Claude 3.5 Sonnet & 60.00 & 60.00 & 33.33 & 60.00 & 20.00 & 46.66 & 93.33 & 75.00 & 78.57 & 46.67 & 64.29 & 53.33 & 20.00 & 62.50 \\
Microsoft Phi 4 & 56.00 & 62.00 & 46.66 & 46.66 & 33.33 & 33.33 & 66.66 & 68.75 & 64.29 & 53.33 & 57.14 & 26.67 & 06.67 & 41.67 \\
\bottomrule
\end{tabular}
\vspace{3ex}
\parbox{\linewidth}{\tiny \textbf{Legend:} \textbf{F1}: Business Math \& Stats; \textbf{F2}: Business Econ \& BCK; \textbf{I1}: Adv. Accounting; \textbf{I2}: Corp. Laws; \textbf{I3}: Taxation; \textbf{I4}: Cost \& Mgmt. Acct.; \textbf{I5}: Auditing \& Ethics; \textbf{I6}: Fin. \& Strat. Mgmt.; \textbf{FN1}: Fin. Reporting; \textbf{FN2}: Adv. Fin. Mgmt.; \textbf{FN3}: Adv. Auditing; \textbf{FN4}: Direct Tax Laws; \textbf{FN5}: Indirect Tax Laws; \textbf{FN6}: Integrated Business Sol.; \textbf{Inst}: Instruct}
\end{table}

\subsection{Exam Level-wise Breakdown}

Table~\ref{tab:llm_levels} displays the accuracies of various models evaluated across the three CA-Ben levels. As per observations, CA-ThinkFlow acquires the highest accuracy score at the Foundation level (80.39\%), and also outperforms almost all the given state-of-the-art models at the Final level (except GPT-4o and Claude-3.5-Sonnet). It also outperforms Mistral Large and the base 14B-Deepseek-R1 model, validating the impact of the RAG mechanism implemented. This validates the strong reasoning capability and robustness of CA-ThinkFlow across all difficulty levels in comparison to significantly larger and computationally inefficient models like Claude 3.5 Sonnet, GPT-4o, LLaMA-3.1-405B-Instruct, and others, especially at the Foundation and Final levels.

\begin{table}[ht]
    \centering
    \caption{Cumulative accuracy (\%) of LLMs across CA-Ben levels.}
    \label{tab:llm_levels}
    \begin{tabular}{lccc}
        \toprule
        \textbf{Model} & \textbf{Foundation} & \textbf{Intermediate} & \textbf{Final} \\
        \midrule
        \rowcolor{green!20} 
        \textbf{CA-ThinkFlow} & \textbf{80.39} & 48.05 & 48.63 \\
        \rowcolor{gray!25}   
        14B-Deepseek-R1 & 58.73 & 34.93 &	32.97 \\
        GPT-4o & 54.00 & 53.61 & \textbf{55.28} \\
        LLaMA-3.3-70B-Instruct & 57.50 & 53.61 & 41.65 \\
        LLaMA-3.1-405B-Instruct & 56.00 & 49.37 & 44.01 \\
        Mistral-Large & 48.50 & 46.59 & 38.47 \\
        Claude-3.5-Sonnet & 60.00 & \textbf{54.72} & 54.23 \\
        Microsoft-Phi-4 & 59.00 & 49.23 & 41.63 \\
        \bottomrule
    \end{tabular}
\end{table}

\subsection{Subject-wise Breakdown}
\begin{figure*}[ht]
\centering
\includegraphics[width=\linewidth]{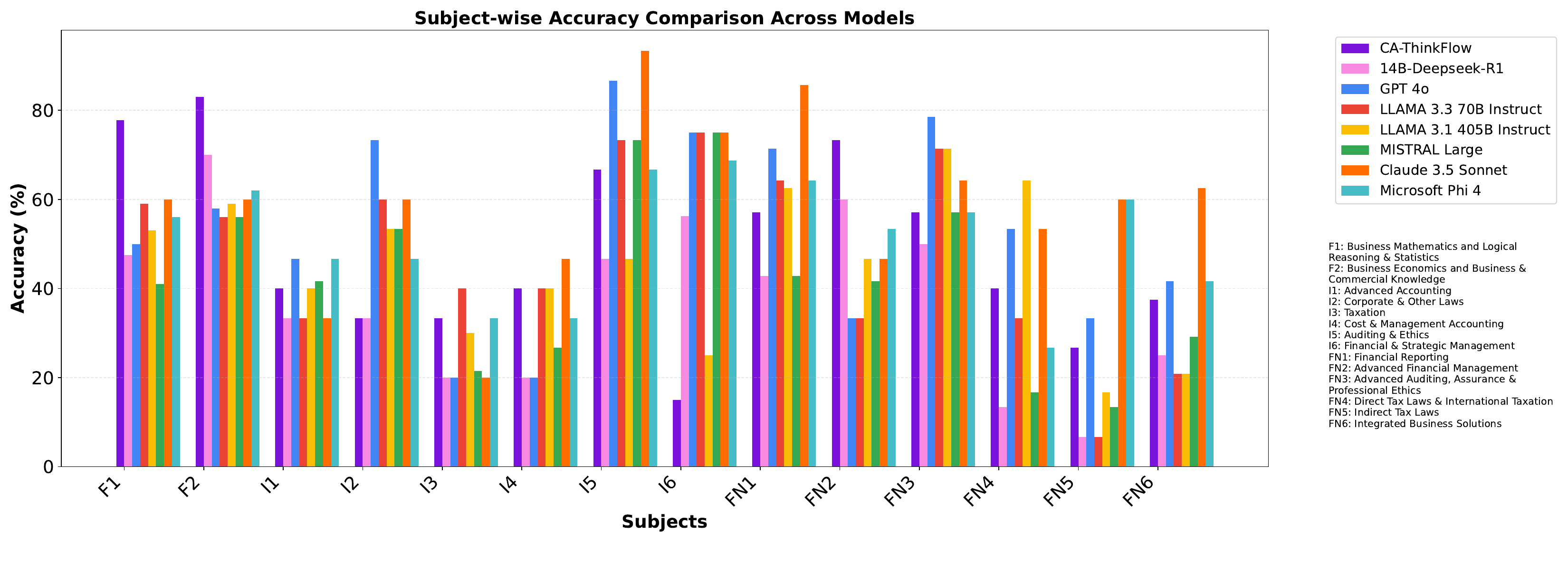}
\caption{Subject-wise accuracy comparison of all evaluated models across 14 CA-Ben domains (F1--FN6). Each bar represents the model’s accuracy for a specific subject, color-coded by model.}
\label{fig:bar_subjectwise}
\end{figure*}

Figure~\ref{fig:bar_subjectwise} represents the accuracy of each model that was evaluated for each subject across the 14 domains of the CA-Ben benchmark. The bar chart with grouped bars directly and distinctly compares model performance in every subject area of Chartered Accountancy, thus throwing light on the different patterns, such as strengths and weaknesses across topics belonging to the foundational, intermediate, and final levels for all the models.

\subsection{Model-Specific Comparative Analysis}
\begin{figure}[H]
    \centering
    \includegraphics[width=\linewidth]{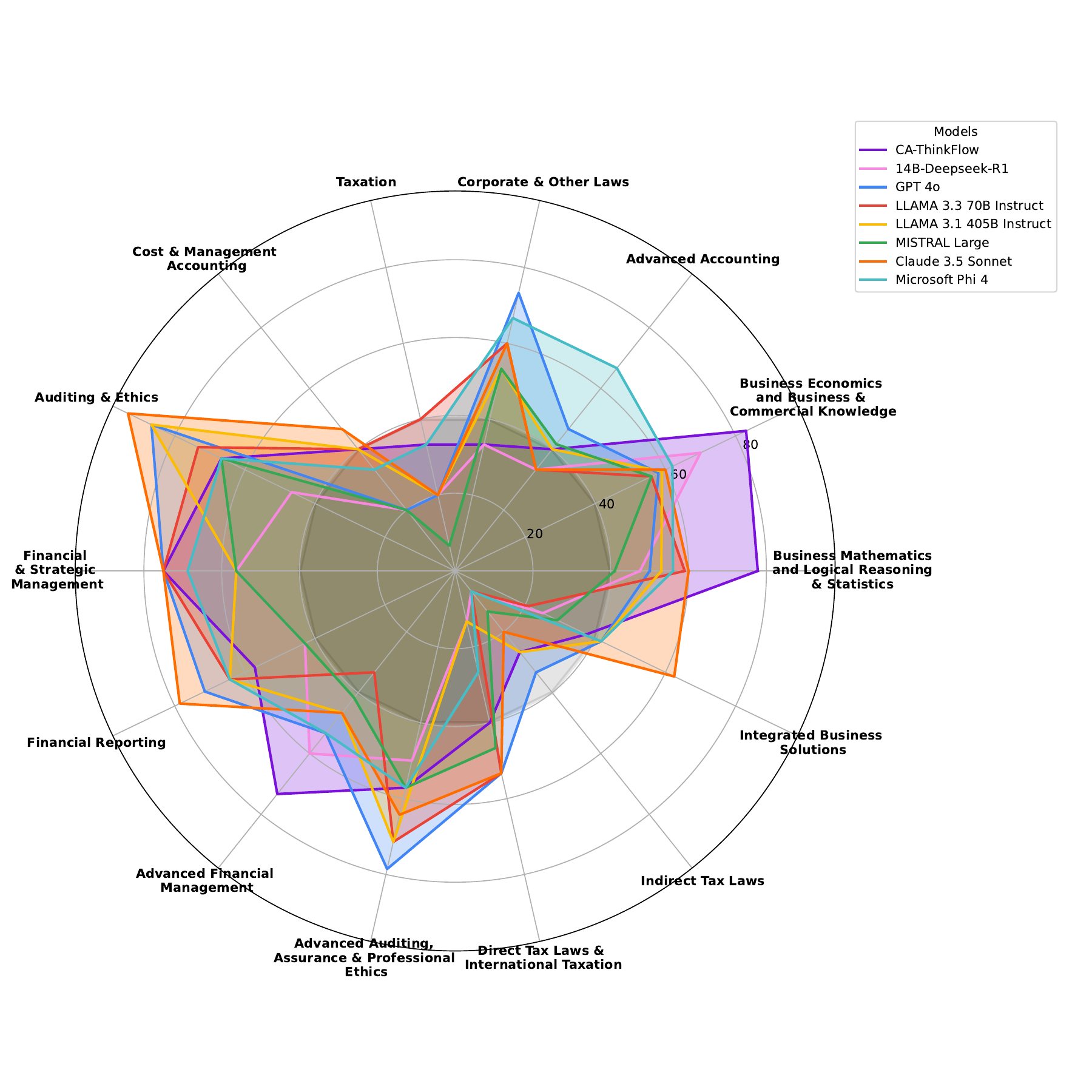} 
    \caption{An overall comparison of models across each individual exam.}
    \label{fig:last}
\end{figure}

CA-ThinkFlow demonstrated unique strengths across different domains in the CA benchmark, reflecting its varied capabilities in greater reasoning and numerical skills in the Foundation Level and comparable scores in Final Level. The radar chart in Figure~\ref{fig:last} clearly visualizes an overall comparison of models across each individual exam. While some models display balanced performance across most domains, CA-ThinkFlow reveals a sharper contrast, excelling in certain areas like Business Economics and Business \& Commercial Knowledge, Business Mathematics and Logical Reasoning \& Statistics, Advanced Financial Management, and others with comparable performance compared to other SOTA models. 

\subsection{Scholastic Reliability Coefficient (SRC) Analysis}

We evaluate the Scholastic Reliability Coefficient (SRC) to assess model consistency across increasing difficulty levels in CA-Ben. Table~\ref{tab:src_scores} shows that CA-ThinkFlow scored an SRC of 68.75\% which matches the performance of top state-of-the-art systems including Claude~3.5~Sonnet and GPT-4o while exceeding other systems. The results show that CA-ThinkFlow operates as a parameter-efficient quantized system which delivers high reliability and consistent performance throughout its foundation and final-level assessments because it possesses strong domain adaptability and reasoning capabilities.

\begin{table}[H]
\centering
\caption{Model rankings by Scholastic Reliability Coefficient (SRC).}
\label{tab:src_scores}
\begin{tabular}{@{}lccccc@{}}
\toprule
\textbf{Model Name} & \shortstack{Foundation Passes} & \shortstack{Intermediate Passes} & \shortstack{Finals Passes} & \shortstack{Weighted Score} & \shortstack{SRC (\%)} \\
\midrule
\rowcolor{green!20}
\textbf{CA-ThinkFlow} & 2/2 & 4/6 & 4/6 & 22 / 32 & \textbf{68.75} \\
\rowcolor{gray!25}
14B-Deepseek-R1 & 2/2 & 2/6 & 3/6 & 15 / 32 & 46.87 \\
GPT-4o & 2/2 & 4/6 & 4/6 & 22 / 32 & \textbf{68.75} \\
LLaMA-3.3-70B-Instruct & 2/2 & 4/6 & 3/6 & 19 / 32 & 59.38 \\
LLaMA-3.1-405B-Instruct & 2/2 & 4/6 & 3/6 & 19 / 32 & 59.38 \\
Mistral-Large & 2/2 & 3/6 & 4/6 & 20 / 32 & 62.50 \\
Claude~3.5~Sonnet & 2/2 & 4/6 & 4/6 & 22 / 32 & \textbf{68.75} \\
Microsoft-Phi-4 & 2/2 & 4/6 & 3/6 & 19 / 32 & 59.38 \\
\bottomrule
\end{tabular}
\end{table}

\subsection{Systemic Bottlenecks}
The original CA-Ben study reported a 100\% failure rate, where no models were able to surpass a 40\% score in a subject. The two subjects of Taxation (I3) and Indirect Tax Laws (FN5) showed these systemic bottlenecks. 
Our system did not clear the Systemic Bottlenecks because the core reasoning abilities of Large Language Models show fundamental knowledge gaps that exist across all domains.

\noindent The results demonstrate that CA-ThinkFlow delivers better parameter efficiency and balanced reasoning performance across foundation and final levels while exceeding the performance of full-precision systems that require more resources.

\section{Conclusion}
\label{sec:conclusion}
The research studies how contextual accuracy and deep understanding work together in the assessment process for chartered accountancy tests. We introduced CA-ThinkFlow as a simplified Retrieval-Augmented Generation (RAG) framework which uses the built-in Chain-of-Thought reasoning of the 4-bit quantized 14B-DeepSeek-R1 model. The system provides high-quality contextual information to the language model which enables it to use its internal reasoning methods for query assessment and synthesis. The multi-level CA-Ben benchmarking of the framework shows that CA-ThinkFlow delivers outstanding parameter efficiency results. The system delivers noteworthy results for all three exam levels while matching the performance of much larger commercial models which operate at full precision and no quantization.

\section{Future Work}
\label{sec:future}
The general performance of CA-ThinkFlow proves to be strong, but its performance needs critical improvement because it cannot overcome essential obstacles which exist in complex fields like Taxation (I3) and Indirect Tax Laws (FN5). Future work should focus on creating a more advanced retrieval pipeline which will enable us to handle both multi-chunk synthesis and dynamic context scaling for our multi-faceted tax query system. It should also investigate domain-specific supervised fine-tuning methods which will enable the base reasoning model to develop deeper regulatory knowledge through examination of Indian financial statutes before the retrieval process. Finally, expanding the system to ingest real-time, evolving tax regulations and incorporating multilingual support will further bridge the gap toward deployable, expert-level financial AI.


\appendix

\section{Data Availability}
\label{app:data_availability}
This study uses official ICAI Study Material as context documents, processed through our pipeline for reproducibility.
\begin{itemize}
    \item Foundation Course: \url{https://www.icai.org/post/foundation-course}
    \item Intermediate Course: \url{https://www.icai.org/post/intermediate-course}
    \item Final Course: \url{https://www.icai.org/post/final-course}
\end{itemize}

For testing data, we used the CA-Ben benchmark under a zero-shot setup, simulating a cold-start scenario~\cite{gupta2025caben}.
\begin{itemize}
    \item Benchmark repository: \url{https://github.com/thejatingupta7/LLMCA}
\end{itemize}

\bibliographystyle{unsrt}  
\bibliography{references}

\end{document}